\documentclass{article}

\usepackage[utf8]{inputenc}
\usepackage{authblk}
\usepackage{setspace}
\usepackage[margin=1.25in]{geometry}
\usepackage{graphicx}
\graphicspath{ {./figures/} }
\usepackage{subcaption}
\usepackage{amsmath}
\usepackage{xcolor}
\usepackage{algorithm}
\usepackage{algorithmic}
\usepackage{siunitx}
\usepackage{ntheorem} 
\theoremseparator{:} 

\newtheorem*{hyp*}{Hypothesis \protect\hypnumber} 

\newcommand{\hypnumber}{}

\usepackage[style=nejm, 
citestyle=numeric-comp,
sorting=none]{biblatex}
\title{Bridging AI Innovation and Healthcare Needs: Lessons Learned from Incorporating Modern NLP at The BC Cancer Registry}

\author[1,2*]{Lovedeep Gondara}
\author[3]{Gregory Arbour}
\author[3]{Raymond Ng}
\author[1,2]{Jonathan Simkin}
\author[1]{Shebnum Devji}

\affil[1]{British Columbia Cancer Registry, Provincial Health Services Authority, Vancouver, Canada.}
\affil[2]{School of Population and Public Health, University of British Columbia, Vancouver, Canada.}
\affil[3]{The Data Science Institute, University of British Columbia, Vancouver, Canada.}
\affil[*]{Address correspondence to: lovedeep.gondara@ubc.ca}

\date{}

\begin{document}

\maketitle


\begin{abstract}
Automating data extraction from clinical documents offers significant potential to improve efficiency in healthcare settings, yet deploying Natural Language Processing (NLP) solutions presents practical challenges. Drawing upon our experience implementing various NLP models for information extraction and classification tasks at the British Columbia Cancer Registry (BCCR), this paper shares key lessons learned throughout the project lifecycle. We emphasize the critical importance of defining problems based on clear business objectives rather than solely technical accuracy, adopting an iterative approach to development, and fostering deep interdisciplinary collaboration and co-design involving domain experts, end-users, and ML specialists from inception. Further insights highlight the need for pragmatic model selection (including hybrid approaches and simpler methods where appropriate), rigorous attention to data quality (representativeness, drift, annotation), robust error mitigation strategies involving human-in-the-loop validation and ongoing audits, and building organizational AI literacy. These practical considerations, generalizable beyond cancer registries, provide guidance for healthcare organizations seeking to successfully implement AI/NLP solutions to enhance data management processes and ultimately improve patient care and public health outcomes.

\end{abstract}

\newpage

\section{Introduction}
Cancer registries are vital in healthcare systems, providing critical data for cancer surveillance, research, and public health policies \cite{parkin2006evolution}. However, the manual processes involved in data extraction from pathology reports and clinical notes are often labor-intensive, time-consuming, and prone to inconsistencies, leading to backlogs and potential delays in data availability \cite{santos2022automatic}. With the recent successes in machine learning (ML) \cite{panesar2019machine}, especially in Natural Language Processing (NLP) \cite{roy2021application}, Language models (LMs) offer a promising foundation to automate and enhance these processes.

Our team has decades of experience in ML and NLP. For the last four years, we have been working on training and deploying various NLP models at British Columbia Cancer Registry (BCCR) for a wide array of tasks, including report classification, information extraction, and text segmentation. Our work primarily covers pathology reports but also includes imaging documents such as CT and PET scan modalities. We have published multiple papers focused on this area, such as for tumor reportability classification \cite{gondara2024classifying}, tumor group classification \cite{gondara2025elm}, relapse detection \cite{lee2024automated}, report segmentation \cite{fung2025using}, auditing language models in a production setting \cite{gondara2025clinical}, etc.

Even with the extensive experience, we faced several challenges bringing the solutions from inception to production and deployment. This paper presents lessons learned from our experience implementing different NLP techniques to automate information extraction from pathology reports at BCCR. We aim to provide practical insights for healthcare organizations considering similar deployments, focusing on key considerations across the project lifecycle. Our findings are not limited to cancer registries, but are generalizable across the healthcare landscape, where we emphasize the crucial role of clear problem definition, taking an iterative approach to the final solutions, interdisciplinary co-design and collaboration, and a pragmatic approach to model selection and deployment. While NLP is technically a subfield of AI focusing on text and language data, for the purposes of this discussion on applying various NLP techniques in healthcare, the terms ``AI", ``ML" and ``NLP" are often used interchangeably.

\section{The Project Goals and Milestones}
The success of any applied AI project hinges on a clearly defined problem statement and well-articulated project goals and milestones. In settings where many people from different backgrounds, roles and organizations are involved, aligning all stakeholders on the project as early as possible to have a common understanding is paramount. 
\subsection{Success Beyond Accuracy: Focus on Business Goals}
It is crucial to recognize that the ultimate goal of implementing automated NLP approaches in a healthcare setting is not solely to maximize model accuracy, but to achieve specific business objectives such as improving efficiency, reducing costs, or enhancing patient care. For instance, in a cancer registry context, the initial, poorly defined goal might be:

\emph{Create a NLP solution that is at least 99\% accurate for identifying reportable tumors.}

This solely focuses on accuracy and lacks specificity. A more effective problem statement, aligned with business needs, would be:

\emph{Address the existing two-year backlog in pathology coding, a labor-intensive and expensive process, by automating a method to achieve accuracy similar to current processes in an automated pipeline for reportable tumor identification in pathology reports.}

This revised example clearly articulates the motivation (backlog reduction), the desired outcome tied to reality (accuracy similar to current processes), and the specific task (reportable tumor identification). Such goals are best established when the end-users (like the cancer registry staff in the example above) and the ML experts collaborate to understand the limitations of current methods and propose a viable target for the ML system. A real example from BCCR is that the discussions between subject matter experts and ML experts lead to the creation of an NLP solution that minimizes false positives while keeping false negatives below 2\%. This shows that strategic alignment of projects is fundamental to realizing business value and achieving desired outcomes \cite{vayyavur2024ai}.

Furthermore, ML experts measure ``model" success by false negatives, false positives, area under the receiver operating characteristic (ROC) curve, etc. However, to a business organization, the success of a ``solution" based on a ``model" can be measured by other metrics such as the amount of waiting time to complete the task, and/or the amount of actual time to complete the task. For waiting time, an AI tool can operate 24/7, while human-centric workflows may be interrupted by staff holidays, shortage, etc. For completion time, an NLP solution may significantly reduce the amount of time taken, even if human post-processing is needed. 

Let us return to the BCCR reportable tumor identification example. Let us say that for the sake of illustration, for every 1000 true positive pathology reports, the current system adds 400 false positive reports. Thus, tumor registrars would need to review 1400 reports to filter out the false positives. With an NLP tool, let us say that the false positives are reduced to 100 reports, and the tumor registrars would need to review 1100 reports. But the gain is more because the NLP tool marks up the sentences on why the report is flagged as reportable. Thus, a tumor registrar instead of spending, say, 1 minute to read one report, may now only need to spend 30 seconds on it. In other words, without the tool, human experts spend 1400 minutes to process 1400 reports; with the NLP tool, this task is now completed in $\dfrac{1100}{2}$ = $550$ minutes. Thus, to BCCR, while reducing the false positive rate from 40\% to 10\% is great, the time reduction from 1400 minutes to 550 minutes is even more valuable.

\subsection{Inclusion/Exclusion Criteria and Outcome Variables}
A well-defined problem statement needs to be clear regarding inclusion and exclusion criteria for data, as well as a precise definition of the outcome variable. For example, when extracting information from pathology reports, specifying the types of reports included (example: breast cancer, invasive surgeries only, for specific date ranges) and the exact entities to be extracted (example: tumor grade, stage, biomarkers) is essential. This specificity directly informs data curation, model training, and performance evaluation.

Furthermore, the choice of evaluation metrics should directly reflect the project goals. For instance, if the aim is to minimize false negatives for reportable tumor classification, recall might be prioritized over overall accuracy. In cases where achieving the initial ambitious goal proves challenging, a pragmatic approach involves exploring pared-down versions of the goal that can still deliver significant value. The team need to be engaged throughout the whole process to interpret the intermediate results and, if necessary, revise goals and milestones to stay focused on getting value.  For example, if we are interested in extraction of multiple items from pathology reports (tumor site, tumor histology, etc.), and automated extraction of all items yields lower than desired accuracy, focusing on a subset where the desired accuracy can be achieved might be a viable alternative. Similarly, automating the processing of pathology reports that can be processed with high accuracy can be prioritized with the reports that are harder to handle by NLP methods being marked for manual review.

\subsection{Prioritizing Early Wins}
Securing organizational buy-in is often a significant hurdle for AI projects, especially in the case where the organization does not have AI expertise. The first few projects are typically the most resource-intensive, involving manual data extraction with additional education for the data abstractors, labelers, and stakeholders; and the development and testing of novel deployment pipelines. Skeptical stakeholders may justifiably question the return on investment.

To mitigate this skepticism, it is crucial to strategically select initial projects or small proof-of-concepts with a high probability of success. Even within a single project, it is often possible to strategically select a sub-project with narrower focus to secure early wins. This strategy builds confidence and demonstrates the tangible benefits of AI solutions. Identifying project champions within the organization who can sponsor and support the project is also vital. Adopting a strategy of under-promising and over-delivering can enhance stakeholder satisfaction and build momentum for future AI initiatives \cite{Peters87}.

\section{The Data} 
The quality and representativeness of training data is fundamental to the performance and generalizability of any ML model.
\subsection{Preparing the Training Data}
In supervised learning approaches, labeled datasets, often prepared manually, are necessities. Recognizing the inherent potential for human error in annotation, especially when the data involves complex clinical text, is crucial. Human annotators, even with training, can make mistakes due to task complexity, text ambiguity, typos, and varying interpretations of clinical language. To improve label accuracy, implementing multi annotation, where multiple annotators label the same observation, is a best practice. Thus, developing a ``code book" for annotation is time worth spending. In fact, having multiple annotators to curate a small pilot dataset and compare the annotations may lead to valuable updates to the code book. It is also helpful to maintain a detailed document outlining annotation procedures and examples of common cases as well as edge cases. The document should be updated to incorporate additional cases that are currently not well explained, as the need arises.

Measuring inter-annotator agreement can quantify label reliability and identify areas of ambiguity. If we can only curate a small amount of data and we have a high-stakes scenario, achieving a high level of inter-annotator agreement to ensure the label’s accuracy is crucial. Conversely, if we have enough training data, a single annotator-based approach might work, as machine learning models (including LMs) are known to be robust in presence of noisy labels \cite{agro2023handling}. We can further refine the labels by focusing manual review on cases where the trained model makes incorrect predictions. This error-focused review can efficiently identify mislabeled instances and areas where the model requires further refinement.

\subsection{Training Data Issue: Representativeness}
Training data, especially if obtained through sampling, must be representative of the real-world population on which we intend to apply model inferences. While it may be tempting to use easily available, but potentially less representative data, this can lead to performance degradation in real-world deployment. As highlighted by \cite{bacelar2021monitoring}, sampling bias can significantly impact model fairness and accuracy. Once a training dataset has been constructed, it would be important to conduct due diligence evaluation, against population-level demographic characteristics, of various covariates, such as sex, age, ethnicities, comorbidities, smoking status, and other factors known to be important to the tasks at hand. 

\subsection{Training Data Issue: Drift}
Anticipating and mitigating data drift is another critical consideration. The data distribution may shift over time due to changes in clinical practice, reporting standards, or patient demographics. Therefore, it is essential to consider data drift during project planning. Implementing automated methods to monitor for data drift and establishing procedures for model retraining as needed are crucial for maintaining pipeline performance over time \cite{sahiner2023data}.

\subsection{Training Data Issue: Size}
Determining the minimally sufficient training dataset size is a practical challenge. Larger datasets generally lead to better model performance. However, as discussed later in Section 4.1, data curation is time-consuming and costly. One efficient approach is to start with a smaller dataset and iteratively increase it based on model performance. Factors influencing the required data size include the complexity of the task, the desired performance targets, and the cost of labeling. For example, for a binary task of classifying reportable tumors vs not, assuming a balanced class distribution, we can finetune an NLP model efficiently with a few thousand pathology reports, whereas in case of histology prediction where we have more than ten classes with skewed distribution, we will need a larger training dataset, with tens of thousands of reports.

In case of data scarcity, which is a common challenge in healthcare, several strategies can be used. Techniques such as oversampling minority classes, under-sampling majority classes, and synthetic data generation (example: SMOTE, GANs) can help balance datasets and augment training data \cite{tanaka2019data,wong2016understanding}. Transfer learning \cite{zhuang2020comprehensive} can be especially useful for deep-learning based models. We can also simplify the problem (example: reducing the number of classes) to mitigate the challenges of small training datasets. For cases where we have small, labeled dataset and large unlabeled dataset, we can use automated techniques such as iterative refinement \cite{triguero2014seg} for increasing the labeled dataset size. 

When working with small training datasets, rigorous cross-validation is crucial for robust model evaluation \cite{raschka2018model}. Limiting the number of experiments and focusing on well-validated techniques helps avoid overfitting and ensures better generalization.

\section{The Language Model(s)}
The landscape of pre-trained LMs is vast, ranging from general-purpose models to domain-specific and large-scale to smaller variants. Choosing the appropriate base model requires careful consideration of project goals, computational resources, and data privacy.

\subsection{Domain-Specific Pretraining and Model Size}
While large language models (LLMs) like Llama \cite{touvron2023llama} offer state-of-the-art performance in many NLP tasks, smaller models like BERT \cite{devlin2019bert} can be more computationally efficient and suitable for resource-constrained environments. Furthermore, pretraining models on clinical text can significantly enhance performance in healthcare applications by enabling them to better capture domain-specific language nuances \cite{alsentzer2019publicly}. The trade-off between model size and computational cost must be carefully evaluated, considering the available compute budget and the need for iterative experimentation. Developing custom pretrained NLP models for finetuning, even smaller variants of BERT or RoBERTa \cite{liu2019roberta} can offer significant advantages for domain-specific applications like cancer registries compared to finetuning a generic pretrained model or using a LLM in zero shot capacity.

\subsection{Generative Models: Opportunities and Cautions}
Generative models, including LLMs, offer exciting possibilities in healthcare. However, their application requires careful consideration of online vs. offline uses, error management, and ethical implications. Generative models can be used in online settings (example: API calls for a cloud-hosted LLM) or offline settings (example: on-premise, air-gapped). Online uses can be budget friendly and convenient. However, they require deep understanding and thorough analysis of the data flow (to and from the cloud) and stringent privacy compliance, while offline uses allow for more flexibility at the expense of larger initial investment. In healthcare, hosting and training models locally may be essential to comply with privacy regulations \cite{edemekong2018health} and organizational policies.

Generative models are prone to ``hallucinations", generating plausible but factually incorrect information \cite{huang2025survey}. In healthcare, this is a critical concern. Some strategies to minimize hallucinations include careful prompt engineering, restricting the model to extractive summaries, i.e., only highlighting actual sentences in the documents, Retrieval-Augmented Generation (RAG), output verification, etc. \cite{li2024enhancing,liu2024exploring}.

\subsection{Selecting the Best Model}
It is tempting to gravitate towards the ``trendiest" and most powerful models. However, a pragmatic approach involves evaluating a variety of models, including simpler methods, to compare and identify the most effective solution for the specific use case. No single model is universally optimal. Taking time to evaluate the strengths and weaknesses of candidate methods and testing a range of options is crucial. There are scenarios where a simple, regular expressions based approach is more appropriate than the latest LLM.

\subsection{Hybrid Approaches}
Often, the most effective solutions involve combining different NLP techniques, leveraging their respective strengths. A hybrid approach might involve using regular expressions for straightforward cases, question-answering based model for relevant information extraction, classification models for moderately complex instances, and LLMs for more nuanced or ambiguous reports. Cases that remain challenging can be flagged for manual expert review, creating a human-in-the-loop system.

Understanding the strengths and limitations of each approach is crucial. Regular expressions work well for pattern-based extraction, eliminate hallucinations, but lack contextual understanding and susceptible to format shifts. Classifiers are efficient but require labeled data and may struggle with limited training examples. LLMs are powerful but can be computationally expensive and require careful prompting and validation. 

Employing ensemble models, parallel or sequential, where we aggregate predictions from multiple diverse models, can improve prediction robustness and reduce error rates, for classification and generative tasks. Ensembles are most effective when individual models are diverse in architecture, training data, or preprocessing \cite{ganaie2022ensemble}.

Beyond combining modeling techniques, addressing the structure of the input document itself is often critical, especially for long texts containing irrelevant information. We refer to this process as report segmentation. For instance, pathology reports frequently include sections on clinical history, gross descriptions, addendums and procedural details that might be irrelevant to a specific task like identifying tumor histology. It is worthwhile to invest in methods to properly segment documents using techniques ranging from regular expressions to specialized question-answering models \cite{fung2025using}, so as to isolate only the text useful to the modeling problem at hand. Effective segmentation can substantially simplify downstream NLP tasks, whether they employ a single model or a complex hybrid system.

\subsection{Data Privacy}
Protecting patient data privacy is paramount in healthcare. The increasing application of small and large language models introduces significant privacy risks, as these models have demonstrated a propensity to inadvertently memorize and potentially leak portions of their training data \cite{carlini2021extracting}. Adversaries can exploit this vulnerability by strategically querying a trained model to extract sensitive training examples, sometimes verbatim \cite{nasr2023scalable}. Hence, it is crucial to integrate privacy into the development lifecycle before training language models on sensitive healthcare datasets. Two primary approaches for mitigating these risks are differential privacy (DP) \cite{dwork2014algorithmic} and data anonymization. DP offers mathematically rigorous, provable guarantees against certain types of leakage but often introduces noise that can degrade model performance. Conversely, anonymization techniques aim to obscure or remove identifiable information, potentially maintaining higher model utility. However, they typically lack strong theoretical guarantees and can be susceptible to re-identification attacks \cite{sweeney2002k}. The selection between these methods, or potentially a hybrid approach, necessitates a careful evaluation, balancing the required level of privacy guarantee against the acceptable impact on model performance.

\section{Error Mitigation} 
Despite advancements in NLP, errors are not completely eliminated. Strategies for minimizing and managing errors are crucial for building trust and ensuring safe deployment of AI methods in healthcare.

\subsection{Human-in-the-Loop and Audits}
Given the inherent limitations of model interpretability \cite{singh2024rethinking}, a human-in-the-loop approach, where some cases are reviewed by human subject matter experts (SMEs), is often beneficial. Cases for review can be selected based on model uncertainty, low confidence predictions, or the criticality of accurate prediction.

The true measure of model success is its performance in the real-world production environment, not just on training or test sets. Test set performance is merely a proxy for production performance. Factors such as overfitting, differences between training/test data and deployment data, and data drifts can lead to significant performance discrepancies in production. Therefore, rigorous validation in the production environment is essential to ensure reliability. This is particularly critical if the model is intended for critical domains such as healthcare.

Before deploying models into production, conducting a small-scale audit of model predictions with SMEs can help identify and rectify obvious errors or biases. Regular auditing of operational models, such as every six months, is crucial to monitor performance, detect data drift, and ensure ongoing effectiveness. This raises the question on how to select the data for audit, and how many samples we need for the audit. Using methods grounded in theory, such as \cite{gondara2025clinical}, can provide a structured framework for auditing and maintaining model performance over time. In fact, we believe that one key benefit of using AI is the opportunities to provide continuous enhancements to the models based on regular audits and error analyses. 

\section{Importance of Collaboration and AI Literacy}
Successful implementation of AI in healthcare requires not only technical expertise, but also a workforce or a user base with sufficient AI literacy \cite{ng2021ai,chetty2023ai}.

\subsection{Engaging Domain Experts}
Developing AI tools for healthcare demands a strong collaboration between clinical, machine learning and NLP experts \cite{bertagnolli2023advancing,nasarian2024designing}. Clinical experts ensure that the AI system addresses genuine medical needs, aligns with clinical standards, and adheres to ethical guidelines. ML and NLP experts contribute technical knowledge in model development, data processing, and algorithm design. Regular communication and feedback between these groups is vital for catching errors early, ensuring alignment with clinical goals, and facilitating iterative improvements.

Ignoring either clinical or ML expertise can lead to flawed development and implementation, potentially resulting in operational inefficiencies or, more seriously, patient harm. A well-coordinated, multidisciplinary approach is essential for developing safe, effective, and trustworthy AI systems in healthcare. From our own experiences, the engagement better starts from the early onset of the project, not when the AI models or tools are close to completion, which leads to the following recommendation.

\subsection{Engaging Users in co-designing and co-development}
Involving users from the initial design phase allows for aligning system objectives with user workflows and expectations. This co-design approach identifies core functionalities, anticipates challenges, and ensures the developed models address practical needs. Continuous user feedback loops during development enable iterative refinements and enhance usability. User testing in real-world settings provides invaluable insights into performance, reliability, and potential blind spots. Extending user involvement into auditing and maintenance phases is crucial for monitoring performance, identifying data drift, and ensuring ongoing alignment with evolving user needs.

\subsection{Building AI Literacy}
AI literacy, defined as the ability to understand and communicate AI concepts, is crucial for all team members involved in the project. Educating stakeholders about AI methods, data preprocessing, model training, and validation fosters transparency, trust, and informed decision-making \cite{ng2021ai,chetty2023ai,cetindamar2022explicating}. Understanding AI fundamentals helps users critically evaluate AI-driven outputs and recognize potential biases, mitigating the risk of perceived ``black box" decision making.

AI literacy promotes a collaborative environment where data annotators, SMEs, and technical teams can effectively communicate and contribute. When annotators understand the impact of their labels, and SMEs grasp the importance of data quality, they are more likely to take ownership and adhere to rigorous standards \cite{schmer2024annotator}. Further, their participation also helps decrease fear (of replacement or new technology) and increases confidence. This inclusive approach improves data quality, boosts team engagement, and ultimately leads to robust, trusted and ethical AI systems. 

One common problem at the beginning of a project is that domain experts may feel that they do not have sufficient AI literacy to be involved. ML experts should avoid excessive uses of technical jargons and try to communicate the essence of the concepts. In our experience, allowing domain experts to pick up on AI concepts through immersion, while taking time, is valuable for long-term success in team communication. 

\section{Build vs. Buy: Considerations for Off-the-Shelf AI Tools}
While this paper primarily focuses on building custom AI solutions, many healthcare organizations may consider purchasing off-the-shelf AI tools, especially for the scenario where there is a lack of in-house expertise. This approach presents both potential benefits and unique challenges.

Off-the-shelf tools can offer faster deployment and reduced upfront development costs. However, they may lack customization for specific needs and goals, may not be transparent in their underlying methods, and can create vendor lock-in. Furthermore, their performance may not be adequately validated for the specific context within a healthcare system. 

In the following, we review the challenges of buying off-the-shelf tools based on the various criteria discussed so far.

\begin{itemize}
    \item \textbf{Business goals.} Off-the-shelf tools are developed with specific goals in mind, particularly catering to the perceived market. The problem that the tool tries to solve may overlap with, but not necessarily identical to the exact problem that the organization needs to solve. A strong overlap may still not be sufficient because additional effort may be needed to retrofit the off-the-shelf tool to the workflow of the organization. This may result in adding more work to the intended users, rather than reducing their workload.
    
    \item \textbf{Training data representativeness.} A common problem with off-the-shelf tools is that there is no explicit description of the size and representativeness of the training data. Training data bias need to be carefully identified and evaluated. For example, an off-the-shelf tool may not have sufficient data for certain desired age groups, ethnicities, and comorbidities, leading to fairness concerns. 
    
    \item \textbf{Training data drifts.} It is important to understand when there is data drift, whether an off-the-shelf tool can be retrained to accommodate the drift. A ``frozen" model over time may become stale, and less and less effective, unless continuous learning can be applied. Furthermore, the organization may need to factor this ongoing retraining, or more generally maintenance cost in its initial decision to purchase the off-the-shelf tool or not.
    
	\item \textbf{Choices of LLM.} Continuing on the topic of data evolution, the past few years have witnessed LLMs evolving at breakneck speed. It is important to understand whether, or how easy, the off-the-shelf tool can be seamlessly upgraded to a more powerful version of an LLM, when one is available. For example, minimizing hallucinations from LLMs is a hot topic of research and development. Thus, an organization may be keen to enhance the off-the-shelf tool with a new LLM that is more effective in dealing with hallucinations.
    
    \item \textbf{Error mitigation.} This issue is quite related to the earlier discussion on training data drifts. The key point here is to make sure that the off-the-shelf tool allows easy audits to be done, and provides useful information to help error analyses and continuous learning.
    
    \item \textbf{Engaging domain experts.} For an off-the-shelf tool, there are few opportunities for co-designing and co-developing. However, it is still important to co-evaluate the tool with domain experts. It is important to evaluate whether the tool is truly helping to increase productivity and/or providing better decision making and care.
\end{itemize}

To summarize this discussion on off-the-shelf tools, we offer the following list of recommendations, called \textbf{DARE}.

\begin{itemize}
    \item \textbf{D}emand robust validation using data similar to the organization’s own data and inquire about the tool's underlying methodology and error handling.
\item \textbf{A}ssess the tool's flexibility for customization and integration with existing in-house systems.
\item \textbf{R}igorously evaluate internal data compatibility with the tool to assess accuracy, bias, and fairness.
\item \textbf{E}ase of evaluation, related to data security, data privacy, vendor support, and long-term viability.
\end{itemize}

\section{Conclusion}
Leveraging AI in a healthcare setting offers significant potential for improving efficiency and data quality. However, successful deployment requires a pragmatic, iterative, and interdisciplinary approach. In this paper, we have presented the lessons learned from the trenches while working on multiple AI projects within the British Columbia Cancer Registry. Specifically, we have emphasized the importance of clear problem definition and business alignment, prioritizing early wins to build momentum, focusing on data quality and representativeness, adopting hybrid NLP approaches and ensemble methods, ensuring rigorous validation and ongoing monitoring, cultivating AI literacy and user-centric design, and leveraging domain expertise and ethical considerations. By addressing these key points, healthcare organizations can effectively harness the power of AI to enhance critical healthcare data management processes, ultimately contributing to improved patient care, public health outcomes and healthcare efficiency.

\printbibliography

\end{document}